\documentclass[10pt, a4paper]{article}

\usepackage[final]{lrec2026} 
\usepackage{CJKutf8}
\usepackage{times}
\usepackage{latexsym}
\usepackage{hyperref}
\usepackage{graphicx}
\usepackage{subcaption}
\usepackage[T1]{fontenc}
\usepackage{booktabs}

\usepackage[utf8]{inputenc}

\usepackage{microtype}

\usepackage{inconsolata}

\usepackage{graphicx}
\usepackage[dvipsnames]{xcolor}

\title{Real-Time Generation of Game Video Commentary with Multimodal LLMs: Pause-Aware Decoding Approaches}

\name{Anum Afzal$^{1,2}$, Yuki Saito$^{3}$, Hiroya Takamura$^{2}$, Katsuhito Sudoh$^{4}$, \\
{\bf \large Shinnosuke Takamichi$^{5}$, Graham Neubig$^{6}$, Florian Matthes$^{1}$, Tatsuya Ishigaki$^{2}$}}
\address{$^1$School of CIT, Technical University of Munich, \\$^2$National Institute of Advanced Industrial Science and Technology (AIST) \\$^3$The University of Tokyo, $^4$Nara Women's University, $^5$Keio University $^6$Carnegie Mellon University\\
         \{anum.afzal, matthes\}@tum.de, \{takamura.hiroya, 	ishigaki.tatsuya\}@aist.go.jp, gneubig@cs.cmu.edu,\\ yuuki\_saito@ipc.i.u-tokyo.ac.jp, shinnosuke\_takamichi@keio.jp, sudoh@ics.nara\-wu.ac.jp }

\abstract{
Real-time video commentary generation provides textual descriptions of ongoing events in videos.
It supports accessibility and engagement in domains such as sports, esports, and livestreaming.
Commentary generation involves two essential decisions: \textit{what} to say and \textit{when} to say it.
While recent prompting-based approaches using multimodal large language models (MLLMs) have shown strong performance in content generation, they largely ignore the timing aspect.
We investigate whether in-context prompting alone can support real-time commentary generation that is both semantically relevant and well-timed. We propose two prompting-based decoding strategies: 1) a fixed-interval approach, and 2) a novel dynamic interval-based decoding approach that adjusts the next prediction timing based on the estimated duration of the previous utterance. Both methods enable pause-aware generation without any fine-tuning. 
Experiments on Japanese and English datasets of racing and fighting games show that the dynamic interval-based decoding can generate commentary more closely aligned with human utterance timing and content using prompting alone. We release a multilingual benchmark dataset and implementations to support future research on real-time video commentary generation.
 \\ \newline  }

\begin{document}

\maketitleabstract

\section{Introduction}

Real-time video commentary enhances the viewing experience by providing descriptions of ongoing events in the video. Generated commentary can be shown as subtitles or synthesized as speech, supporting accessibility and enriching videos without professional narration—making them more understandable and engaging, especially for non-expert audiences. Traditional automatic commentary generation often relies on supervised pipelines that divide the task into two components: one model determines whether to speak, and another generates the utterance~\cite{Ishigaki2021}.

Recent work using large multimodal language models (MLLMs) opens new possibilities for flexible, domain-adaptable commentary generation~\cite{liu2023visual, openai2023gpt4} without training efforts. However, most prior studies overlook the challenge of \textit{when} to speak, assuming fixed-length video inputs and generating a single utterance per clip. This leaves open the question of whether MLLMs can manage both utterance generation and timing identification via prompting alone.

\begin{figure}
    \centering
    \includegraphics[width=0.95\linewidth]{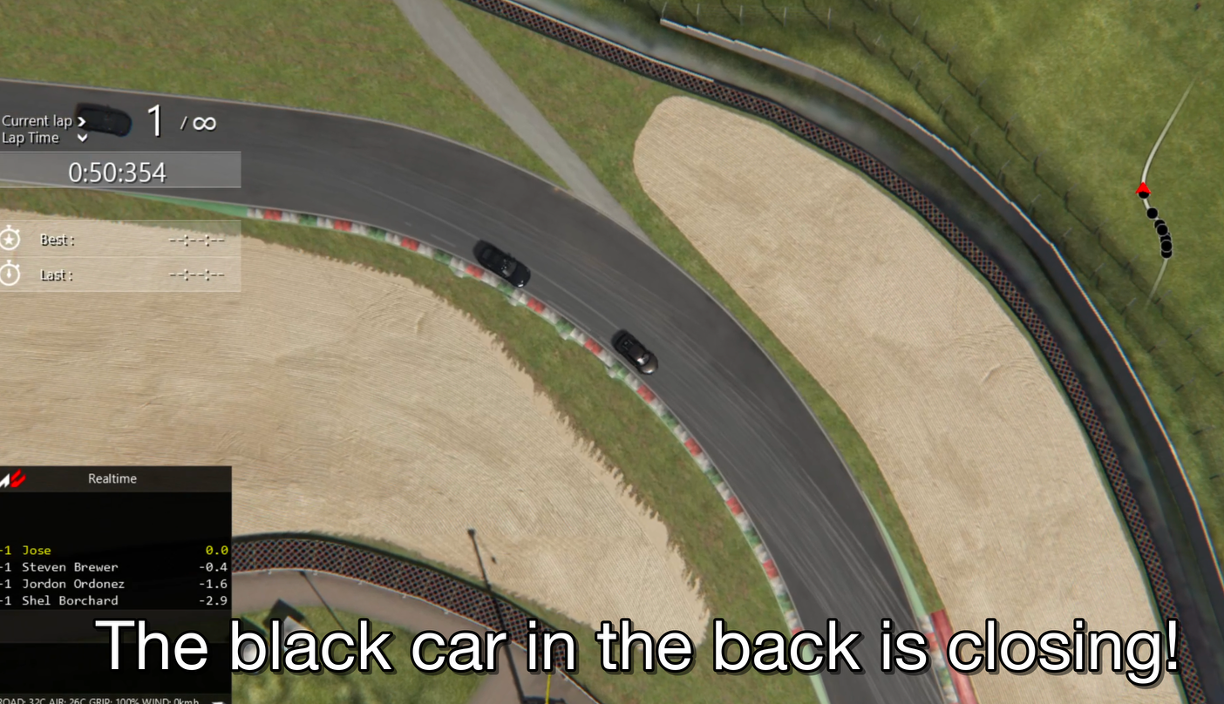}
    \caption{An example of automatically generated commentary for a racing game shown as a subtitle on video.}
    \label{fig:commentary_sample}
\end{figure}

In this work, we address these limitations by exploring prompting strategies that control not only \textit{what} to say, but also \textit{when} to say it. 
Taking inspiration from LLM-based simultaneous machine translation \cite{koshkin-etal-2024-transllama,wang-etal-2024-simultaneous}, we introduce a feedback loop in our decoding strategies, such that the commentary generated in previous steps is passed to the LLM as additional context for deciding whether to speak or remain silent. This feedback loop assists MLLMs in determining if it should WAIT or generate the commentary. Furthermore, using dynamically varying video frame lengths—rather than the fixed-length inputs commonly used in prior work—leads to improved alignment with human utterance, judged by scoring by human annotators, e.g., the average human score is gained from 2.22 to 3.50 in terms of timing appropriateness. These findings demonstrate that MLLMs can exhibit better pause-aware behavior with the dynamic-interval decoding, offering a lightweight alternative to fully supervised or streaming-specific methods.

We investigate two methods: 1) one that queries the model at fixed intervals as a naive extension of existing commentary generators that use fixed-interval of video frames as input, and 2) another that adapts the next prediction time based on prior utterance duration, i.e., by using variable-length input video frames. Our experiments compare these two prompting-based approaches on a multilingual benchmark of racing and fighting games in Japanese and English. We evaluate model outputs based on how closely their utterance timing and content match human commentary.

Our approach is different from existing streaming-based LLMs like LiveCC~\cite{chen2025livecc} and VideoLLM-online~\cite{videollmonline2024}, as they heavily rely on extensive fine-tuning and large amount of labelled data. Additionally, our framework is LLM-agnostic and can be used out of the box by selecting any MLLM of choice. To summarize, our contributions include:

\begin{itemize}
    \item We propose two pause-aware decoding strategies\footnote{Code can be found at \href{https://github.com/anum94/Video2Text}{github.com/anum94/Video2Text/}} for real-time video commentary generation using MLLMs, including a novel feedback-based method that dynamically schedules utterances.
    \item we demonstrate that dynamically adjusting video input intervals improves temporal alignment and semantic relevance without finetuning.
    \item We release a multilingual benchmark across three datasets, enabling standardized evaluation of pause-aware language generation.
    
\end{itemize}

\section{Related Work}

Early approaches to automatic video commentary generation typically divided the task into two subtasks: determining \textit{when} to speak, and generating \textit{what} to say.  
Rule-based systems~\cite{Kim2020,Taniguchi2019,kubo,ishigaki-etal-2023-audio} and supervised classifiers~\cite{Ishigaki2021} were commonly used to detect salient moments in the video, while template-based~\cite{Kim2020} or neural models~\cite{Ishigaki2021,wang-yoshinaga-2024-commentary,someya-etal-2025-live} and extraction method~\cite{kubo} were used to obtain utterances.  
Although effective in constrained settings, these pipelines required task-specific engineering and annotated training data, making them difficult to generalize.

Recent work has explored the use of large language models (LLMs), particularly multimodal LLMs (MLLMs), to simplify commentary generation via prompting.  
Several studies use pre-annotated event timestamps to guide LLM outputs~\cite{li2025soccercomment,mori-etal-2025-live,someya-etal-2025-live}, while SCBench~\cite{ge2024scbench} frames the task as clip-level captioning, prompting the model to produce a single sentence per segment.
These approaches avoid explicit timing prediction but rely on external segmentation and assume fixed-length inputs, thereby addressing only \textit{what} to say, not \textit{when}.

Other lines of work focus on supervised, streaming generation.  
LiveCC~\cite{chen2025livecc} and VideoLLM-online~\cite{videollmonline2024} generate commentary token-by-token from video input, enabled by finetuning models on domain-specific corpora.  
These systems offer low-latency outputs suitable for real-time applications, but require extensive training and careful calibration.  
Moreover, their streaming nature complicates integration with sentence-based systems such as subtitle displays or speech synthesis, which require discrete, well-timed utterances.

In contrast, our work explores whether general-purpose MLLMs can handle both \textit{what} and \textit{when} to say using prompting alone, without any task-specific fine-tuning.  
Inspired by WAIT/WRITE strategies in simultaneous translation~\cite{kano-etal-2021-simultaneous,koshkin-etal-2024-transllama,wang-etal-2024-simultaneous}, we design pause-aware decoding strategies that simulate real-time generation at the utterance level.  
Unlike prior clip-level prompting or streaming decoding approaches, our methods allow the model to dynamically decide whether to speak or remain silent at each step, using both recent video and past commentary as context.  
This design enables natural pacing and seamless integration with downstream systems, e.g., text-to-speech technologies, while maintaining the flexibility and generality of prompting-based generation.

\section{Real-time Commentary Generation}

\subsection{Problem Formulation}

We consider real-time video commentary generation as a causal sequence generation task. The input is a video stream represented as a sequence of frames $\mathcal{V} = \{v_1, v_2, \dots, v_T\}$, and the output is a sequence of utterances $\mathcal{Y} = \{y_1, y_2, \dots, y_K\}$, where each $y_k$ is either a textual utterance or a special token \texttt{<WAIT>}.
We focus on a real-time setting, where at each decision point $t_i$ (determined by the decoding schedule explained in the following subsections), the model has access to the preceding video frames, i.e., $\mathcal{V}_{\leq t_i} = \{v_1, \dots, v_{t_i}\}$, and a buffer of past utterances $H_{< t_i} = \{y_1, \dots, y_{k-1}\}$, which serve as textual context. Based on this input, it decides whether to generate an utterance or remain silent.
We assume that the generated commentary will be delivered either as subtitles or as audio~\cite{iura2025excitement}, but in our experiments we use subtitles.

\subsection{Decoding Strategies}

We explore whether real-time commentary generation can be achieved using prompting alone, without any finetuning or architectural changes. To this end, we design two prompting-based decoding strategies for MLLMs: a fixed-interval decoding method and a novel dynamic interval-based decoding method inspired by simultaneous translation~\cite{kano-etal-2021-simultaneous,ChoEsipova2016SimulMT}. While the former queries the model at uniform time intervals, the latter adapts the next prediction time based on the estimated duration of the previous utterance.

\subsubsection{Fixed Interval-based Decoding}

\begin{figure*}[t!]
  \centering
  \includegraphics[width=0.90\textwidth]{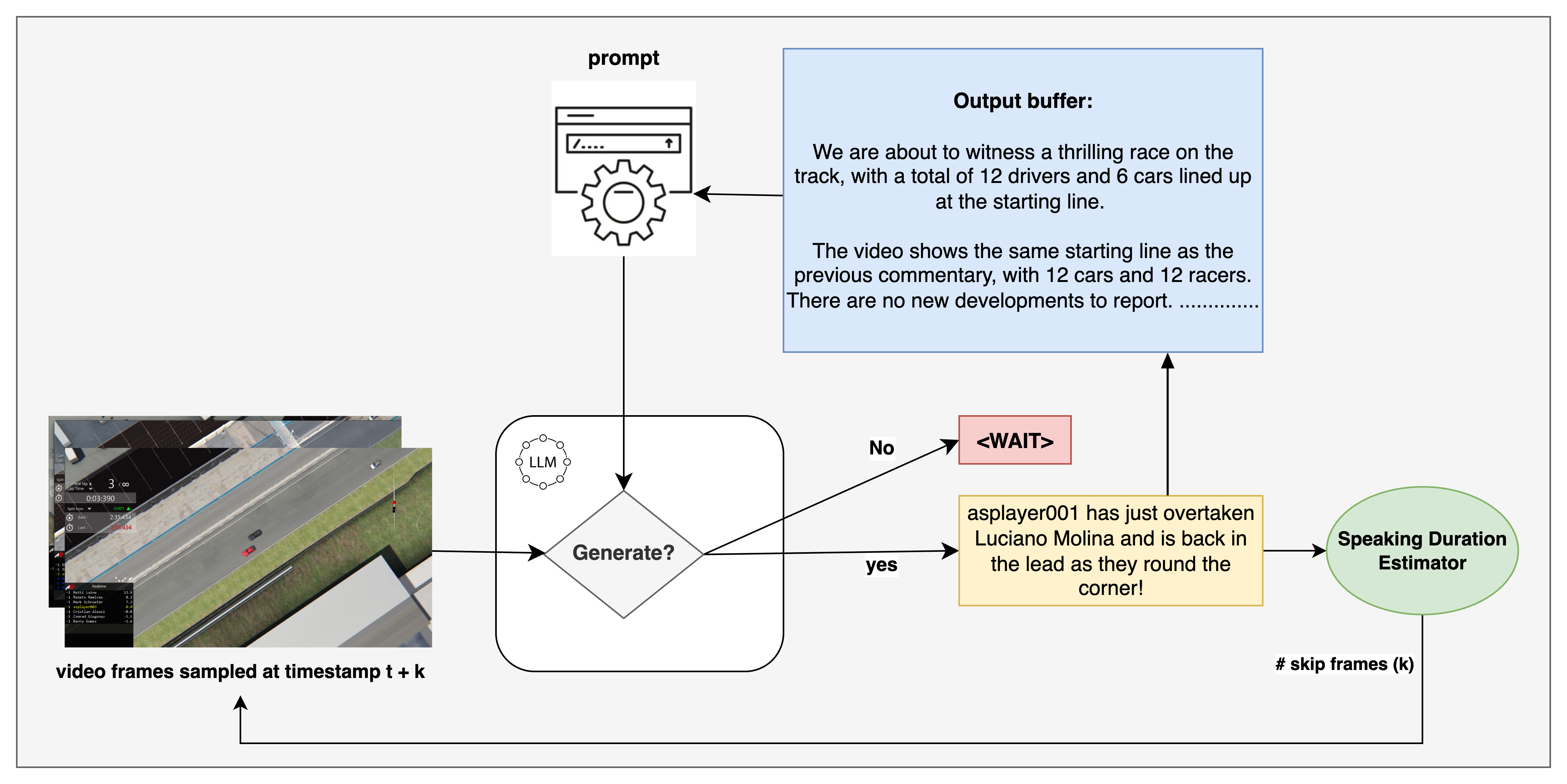}
  \caption{Illustration of both Fixed Interval-based and Dynamic Interval-based decoding strategies queried at uniform intervals of $t$ seconds. For Fixed Interval-based strategy, $k$ is fixed as 0. }
  \label{fig:fixed-interval}
\end{figure*}

The fixed-interval decoding method, as shown in \autoref{fig:fixed-interval} with $k=0$, serves as a straightforward extension of conventional clip-level generation approaches to the real-time setting. Specifically, the model is queried at every fixed interval of $N$ seconds. At each step, it receives a short video clip and is prompted to either generate a textual utterance or output a special \texttt{<WAIT>} token if no update is needed.

\noindent We implemented three variants of Fixed-Interval Decoding; \texttt{Stateless} is a baseline that takes only a video clip, and \texttt{Feedback} takes a list of previously generated utterances in the prompt. \texttt{Feedback (ICL)} is a simple extension that uses demonstrations in the prompt. 

\noindent  While simple, this method inherits limitations from prior work. For example, if $N$ is too short, the system may fail to keep up with real-time inference, especially on resource-constrained hardware. Conversely, if the generated utterance is long but the next query comes shortly after, the model may output another long commentary within a short window. When used as subtitles, this results in rapid consecutive updates that are difficult for users to follow and cognitively process.
These issues highlight the need for a more flexible decoding schedule that accounts for utterance length.
\subsubsection{Dynamic Interval-based Decoding}

To address the limitations of fixed-interval decoding, we propose \textit{dynamic interval-based decoding}, a real-time strategy inspired by WAIT/WRITE policies in simultaneous translation~\cite{kano-etal-2021-simultaneous}. As shown in \autoref{fig:fixed-interval}, the method \texttt{Realtime}, the model dynamically regulates the timing of generating the next utterance based on the duration of previously generated utterances. 

After generating an utterance at time $t_i$, we estimate its speaking duration based on word count and a fixed speech speed rate.
The next prediction time $t_{i+1}$ is then scheduled immediately after this delay. As a result, an LLM is prompted less frequently for longer utterances and more often for shorter ones, which may better align with natural speech rhythm or human reading speed. The estimated speaking duration of each utterance is computed as $\hat{d} = w / r$, where $w$ is the number of words in the generated utterance and $r$ is a fixed speech rate, where $r$ is empirically determined from the validation data in our experiments: 4 words/sec for English and 8 characters/sec for Japanese. In our experiments, the spoken latency is always greater than the latency of MLLM itself, and hence is automatically taken into account. 
Unlike fixed-interval decoding, which operates with a static number of input video frames, feedback-based decoding dynamically adjusts the length of the video segment according to the elapsed time since the last utterance. The motivation of this decoding strategy is to provide the model with sufficient visual context to capture what has changed since the previous utterance—enabling more accurate and context-aware commentary, even without explicit event segmentation.

\section{Experiments}

We empirically evaluate the two decoding strategies, i) fixed-interval and ii) dynamic interval-based decoding, for real-time video commentary generation using MLLMs. All experiments were conducted on a single NVIDIA A100 GPU.

\subsection{Prompts}

We use a unified prompting template across all methods to guide the model in real-time commentary generation. Each prompt consists of three components: 1) a role description (e.g., ``You are a professional commentator''), 2) a list of prior utterances paired with their corresponding video frames, and 3) an instruction to describe the current scene in one sentence (e.g., ``Your task is to generate 1–2 lines of commentary to describe the scene.''). If no significant change has occurred in the video, the model is instructed to output a special token \texttt{<WAIT>} instead of producing a new utterance. This design encourages the model to remain silent when appropriate, enabling pause-aware generation. We use an initialization step at $t_0$ such that the prompt does not include the part containing the previously generated commentary is omitted. The full prompt templates are shown in Appendix~\ref{sec:appendix:prompts}.

\subsection{Datasets}

We evaluate our methods on three datasets in two domains and two languages: car racing (English and Japanese) and fighting games (Japanese). The Japanese racing dataset~\cite{Ishigaki2021} contains commentary aligned with recorded videos of car races. We also use the English version of this dataset. 
An API-based translator is used to create this data, and manual inspection of 100 random examples revealed no translation errors. Manual inspection of 100 random samples revealed no significant translation errors, suggesting that the translation quality is sufficient for experimental use.
The fighting game dataset, SmashCorpus~\cite{saito-etal-2020-smash}, is based on \textit{Super Smash Bros. Ultimate}, with 64 videos annotated with human spoken commentary and corresponding transcripts in Japanese, which we treat as references. Compared to the racing dataset, this corpus features a much faster speaking pace.

These datasets cover varying domains (racing vs. fighting), narrative styles (slow-paced vs. fast-paced), and languages (English and Japanese), enabling comprehensive evaluation of model behavior.
We use 200 randomly sampled racing game videos from the race dataset, each containing approximately 50 utterances on average. We use all videos in SmashCorpus, i.e., 64. All evaluations are conducted on datasets that do not appear in pretraining corpora of these models, making contamination highly unlikely.

\noindent We consider this combined dataset a benchmark for future research on pause-aware, real-time commentary generation. General statistics for the datasets are shown in \autoref{tab:ds_dist}.

\begin{table*}
    \centering
    \begin{tabular}{lrrrrrr}
       
         & \multicolumn{3}{c}{\textbf{Video Durations (sec)}} & \multicolumn{3}{c}{\textbf{Commentary Words (\#)}} \\
          \cmidrule(lr){2-4} \cmidrule(lr){5-7}
        Dataset & avg & min & max & avg & min & max \\
        \toprule      

        Race (en)  & {333.5} & {254}  & {452} & {563.6} & {180} & {866}  \\
        Race (jp)  & {330.8} & {249}  & {365} & {681.0} & {341} & {1128} \\
        Fight (jp)  & {178.2} & {171}  & {196} & {546.1} & {454} & {627} \\

        \toprule
    \end{tabular}
\caption{Average, Minimum, and Maximum length of Videos in seconds,  word count of Reference Commentaries of all 3 datasets }
\label{tab:ds_dist}
\end{table*}

\subsection{Backbone LLMs}

\begin{table*}[h]
\centering
\begin{tabular}{cccccccccc}
\toprule
Model & \multicolumn{3}{c}{Race (en)} & \multicolumn{3}{c}{Race (jp)} & \multicolumn{3}{c}{Fight (jp)} \\ 
\cmidrule(lr){1-1}\cmidrule(lr){2-4} \cmidrule(lr){5-7} \cmidrule(lr){8-10}
 & avg & min & max & avg & min & max& avg & min & max \\
\texttt{Llava~7b} & 1358.3 & 409 & 3238 & 870.9 & 1121.2 & 184& 944.6 & 0 & 1868 \\
\texttt{Qwen~7b} & 2393.9 & 424 & 5309 & 2989.2 & 0 & 6497& 677.2 & 0 & 3432\\
\texttt{GPT-4.1}  & 1241.8 & 166 & 3520 & 1440.0 & 150 & 3563& 1773.9 & 306 & 2615\\
\midrule
\texttt{Reference}  & 563.6 & 180 & 866 & 681.0 & 341 & 1128& 546.1 & 454 & 627\\
\bottomrule
\end{tabular}
\caption{Average words in LLM-generated commentary across all decoding strategies.}
\label{tab:llm-avg-generation-word}
\end{table*}
We compare our proposed dynamic interval-based decoding method (\texttt{Realtime}) with three fixed-interval decoding strategies: \texttt{Stateless}, \texttt{Feedback}, and \texttt{Feedback (ICL)}.  
Following previous work~\cite{Ishigaki2021}, we sample the video at 1 frame per second and use two consecutive frames as input, reflecting the average silence duration of approximately two seconds reported in that study. For \texttt{Feedback (ICL)}, we explored different numbers of in-context examples (2, 4, 6, and 8) in preliminary experiments and found that using 8 demonstrations yielded the best performance. We therefore use 8-shot prompting in all large-scale experiments. The ICL examples are randomly sampled utterance–video pairs from the training set and included directly in the prompt. We use both API-based and open-source models as LLM backbones. As a commercial API-based system, we adopt \texttt{GPT-4.1}. As open-source models, we evaluate \texttt{LLaVA-NeXT-Video} and \texttt{Qwen2.5-VL-Instruct}, both of which support video input and can be used without finetuning.
\subsection{Evaluation Metrics}

We assess two aspects of performance through automatic and qualitative analysis:

\noindent
\textbf{Timing Alignment:} To evaluate how well the model aligns with human timing, we compute the correlation between the timestamps of generated and reference utterances at per-second resolution. In this evaluation, we assume that high-quality commentary not only depends on what is said, but also on saying it at the right time. Therefore, higher correlation indicates better temporal responsiveness and more natural integration of the commentary into the ongoing video. At each time step, we consider the alignment to be a match if both the language model (LLM) and the reference commentary either indicate a pause or both indicate speaking.

\noindent
\textbf{Content Similarity to Human Commentary:} To evaluate whether the model generates commentary that is semantically similar to human references at appropriate times, we compute average ROUGE-L~\cite{lin2004rouge}, and BERTScore~\cite{zhang2019bertscore} scores  between generated and reference utterances for all videos in the test datasets.
Additionally, we segment each video into ten uniform bins (i.e., 10 seconds in our experiments) and compare generated and reference utterances within the same bin. This temporal alignment encourages matching not only in content but also in timing. We compute only BERTScore for each bin since the lexical similarity between generated and reference utterances is very low.
These metrics capture lexical and semantic similarity, reflecting how closely the generated utterance mirrors what a human would say in that moment.


\noindent
\textbf{Human Evaluation}
Apart from the automatic metrics, we also take subjective evaluation into account.
For human evaluation on the English dataset, we asked two human annotators at a rate of 16 euro per hour to watch the videos with the generated commentary as subtitles.
For the Japanese datasets, the evaluation was conducted by a native Japanese speaker with experience in both car racing and gameplay.
We briefly explain the four evaluation criteria, while a detailed version is shown in \autoref{appendix:HE-instructions}:
\begin{itemize}
    \item Including key event identified (KEI) [0-5]: To what degree was the commentary able to identify the transition/changes in the game?
    \item Pause-awareness [0-5]: Was it able to distinguish between key events and stay silent when nothing new happened?
    \item Coherence[0-5]: To what degree did the generated commentary seem coherent?
    \item Naturalness:[0-5] To what degree did the generated commentary seem natural?
\end{itemize}

\begin{table*}[h]
\centering
\resizebox{\textwidth}{!}{
\begin{tabular}{lccccccccc}
\toprule
& \multicolumn{3}{c}{\textbf{Race (English)}} & \multicolumn{3}{c}{\textbf{Race (Japanese)}} & \multicolumn{3}{c}{\textbf{Fighting}}\\
\cmidrule(lr){2-4}\cmidrule(lr){5-7}\cmidrule(lr){8-10}
& Alignment & BERTScore & ROUGE & Alignment & BERTScore & ROUGE & Alignment & BERTScore & ROUGE \\
\midrule
\multicolumn{10}{c}{\textit{LLaVA-NeXT-Video-7B-hf}} \\
\midrule
 
\texttt{stateless} & {0.16} & {0.18}& {5.1}& {0.51} & {0.14}  & {0.1}& {0.18} & {0.28}  & 4.9\\
\texttt{Feedback} & \textbf{0.65}& \textbf{0.20}  & {10.6}& {0.20}& {0.13}  & {0.8}& {0.21}& 0.30& {1.0}\\
\texttt{Feedback (ICL)} & {0.19}& {0.18}& {8.1}& {0.22} & {0.28}  & {5.6}& {0.30}& {0.23}& {0.7}\\
\texttt{Realtime} & {0.25}& {0.19}& {6.4}& {0.19} & {0.23}  & {0.5}& {0.18}& {0.30}& {0.6}\\

\midrule
\multicolumn{10}{c}{\textit{GPT-4.1}} \\
\midrule
\texttt{stateless} & {0.16} & 0.17& {6.2}& {0.15} & {0.25}  & \textbf{7.1}& {0.18}& {0.29}& \textbf{6.3}\\
\texttt{Feedback} & {0.64}& {0.16}  & {9.8}& {0.62} & {0.28}  & {2.5}& {0.20} & {0.28}& {3.2}\\
\texttt{Feedback (ICL)} & \textbf{0.65}& {0.16}  & \textbf{12.1}& \textbf{0.64} & {0.28}  & {2.6}& {0.20} & {0.30}  & {3.8}\\
\texttt{Realtime} & {0.57} & {0.16}  & {10.6}& {0.62} & {0.28}  & {2.5}& {0.27}& \textbf{0.31}& {0.4}\\

\midrule
\multicolumn{10}{c}{\textit{Qwen2.5-VL-7B-Instruct}} \\
\midrule
\texttt{stateless} & {0.16} & 0.18& {6.2}& {0.60} & {0.13}  & {2.5}& {0.78} & {0.13}  & {0.18}\\
\texttt{Feedback} & {0.18}& {0.16}& {4.4}& {0.21} & \textbf{0.29}  & {4.4}& \textbf{0.81}& {0.13}& {2.3}\\
\texttt{Feedback (ICL)} & {0.18}& {0.17}& {6.4}& {0.20} & \textbf{0.29}  & {5.4}& {0.24}& {0.28}& {1.02}\\
\texttt{Realtime} & {0.17}& {0.17}& {4.8}& {0.30} & \textbf{0.29}  & {3.0}& {0.22}& {0.22}& {0.98}\\

\bottomrule

\end{tabular}
}
\caption{Time Alignment [0 - 1], BERTScore F$_1$ [-1 - +1], ROUGE-L (0 - 100) of all models across the three datasets and all four decoding strategies.}
\label{tab:automaticscores}
\end{table*}

\begin{figure}[]
  \centering
  \begin{subfigure}[b]{0.45\textwidth}
    \centering
    \includegraphics[width=\linewidth]{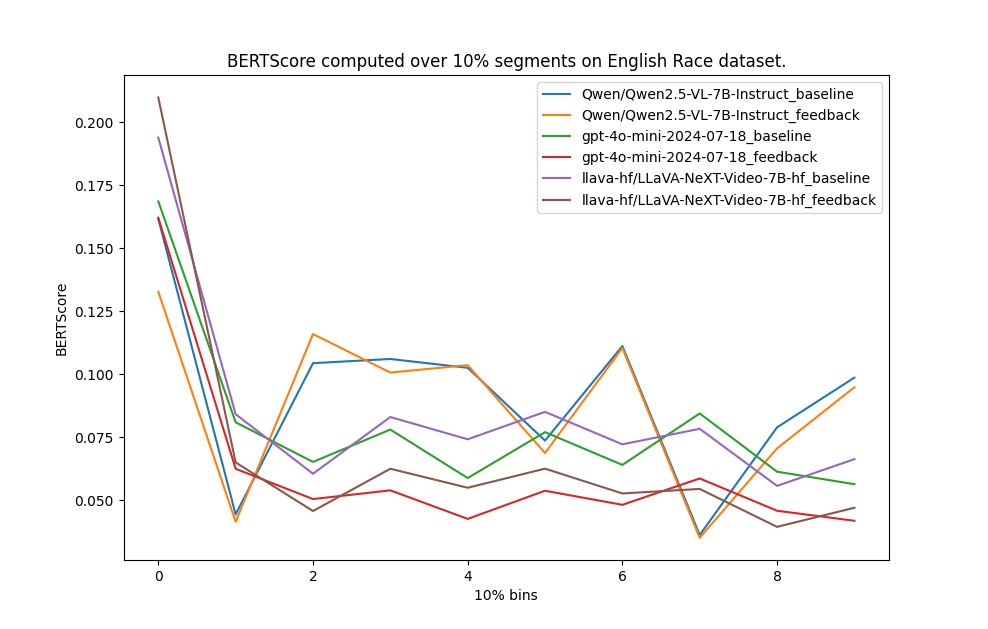}
    \caption{Race (En)}
    \label{fig:img1}
  \end{subfigure}
  \hfill
  \begin{subfigure}[b]{0.45\textwidth}
    \centering
    \includegraphics[width=\linewidth]{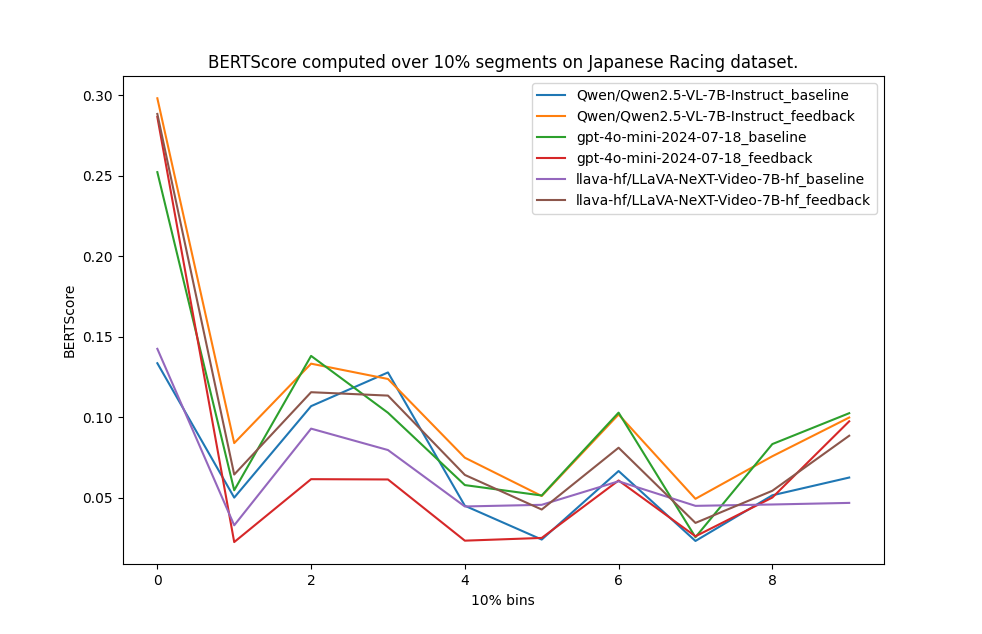}
    \caption{Race (Ja)}
    \label{fig:img2}
  \end{subfigure}
    \hfill
  \begin{subfigure}[b]{0.45\textwidth}
    \centering
    \includegraphics[width=\linewidth]{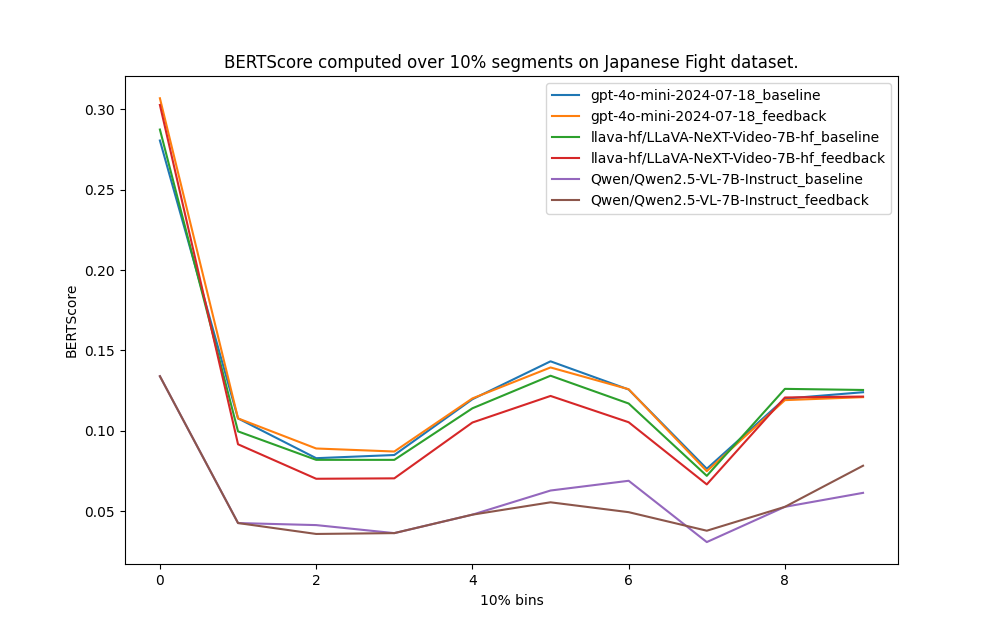}
    \caption{Fighting (Ja)}
    \label{fig:img2}
  \end{subfigure}
  \caption{Contextual similarities between LLM-generated and Reference Commentaries compute over 10\% segments of the whole video.}
  \label{fig:bertscore=bin}
\end{figure}

\section{Results}

\subsection{Automatic Evaluation}
To provide a comprehensive overview of our benchmark, we evaluate performance across three metrics: Alignment (based on generation time), Contextual Similarity via BERTScore, and lexical similarity through ROUGE-L, as presented in \autoref{tab:automaticscores}. In terms of alignment (when to speak?), our experiments demonstrate that the Fixed-Interval decoding approach with ICL examples, outperforms all other methods across all three datasets. Nonetheless, it is important to note that while Llava achieves the same performance as GPT-4.1 on English datasets, its performance deteriorate when evaluated on the Japanese version of the same dataset. Notably, Qwen2.5 shows impressive capabilities when working with Japanese text. Human evaluation reveals that some models, particularly LLaVA-NeXT-Video, often switch between English and Japanese, struggling to maintain consistency in the chosen language during Japanese commentary generation. Additionally, the LLM-generated commentaries are much more verbose than the human commentaries.  This inconsistency, along with the fact that the choice of words vary significantly even for \textit{correct commentaries}, likely contributes to the notably low ROUGE scores. Similarly, BERTScore evaluations in Japanese exhibit considerable variability. These observations suggest that many models still have room for improvement in multilingual capabilities. Due to the language inconsistencies observed in model outputs, we place greater emphasis on subjective evaluation to assess linguistic and contextual coherence.

\subsection{Subjective Evaluation}

As shown in \autoref{tab:subjective_evaluation}, GPT-4.1 consistently outperformed the open-source models across all criteria.  Notably, our proposed \texttt{Realtime} decoding strategy yielded the best or near-best scores in most categories for GPT-4.1, achieving an average of 3.50 in Pause-awareness for Japanese racing commentary and 3.93 in Naturalness for fighting games. These results suggest that dynamic interval-based prompting allows models to more effectively regulate output timing and avoid unnecessary verbosity. Qwen2.5 and LLaVA-NeXT-Video performed competitively in some cases, especially in Japanese commentary settings, but were generally more prone to producing unnatural or mistimed utterances.  
In particular, \texttt{Realtime} decoding improved Pause-awareness by a wide margin over fixed-interval strategies, even for open-source models. These findings validate our hypothesis that prompting-based methods, when combined with timing-aware decoding, can enable pause-aware and contextually fluent commentary without finetuning. The subjective results also demonstrate that such methods are suitable for real-time applications where natural pacing is critical. Although these results do not align with the automatic evaluation scores, they do carry more weight into our evaluation since automatic metrics are known to be unreliable. To provide more insight into both methods, we show parts of commentary generated by both method in \autoref{app:generation_examples}.

\begin{table}[h]
\resizebox{\columnwidth}{!}{
\begin{tabular}{ccccc}
\toprule
step size & stateless & feedback & real-time & avg \\ \midrule
1  &   {0.166}        &     {0.295}     &    {0.693}      &    {0.462}     \\
2  &   {0.163}        &     {0.687}     &    {0.417}      &    {0.421}     \\
5  &   {0.172}        &     {0.202}     &    {0.549}      &    {0.368}     \\
10 &   {0.169}        &     {0.170}     &    {0.460}      &    {0.3151}     \\
\bottomrule
\end{tabular}
}
\caption{Effects of step size on the utterance timing correlation on different decoding strategies}
\label{tab:step_size}
\end{table} 
\begin{table*}[h]
\resizebox{\textwidth}{!}{%
\centering
    \begin{tabular}{lcccc cccc cccc}
    \toprule
     & \multicolumn{4}{c}{\textbf{Race (English)}} & \multicolumn{4}{c}{\textbf{Race (Japanese)}} & \multicolumn{4}{c}{\textbf{Fighting}} \\ 
     \cmidrule(lr){2-5}\cmidrule(lr){6-9}\cmidrule(lr){10-13}
    \multicolumn{1}{c}{\textit{}} & KEI & Pause-aware & Coh & Nat & KEI & Pause-aware & Coh&  Nat & KEI & Pause-aware & Coh& Nat\\
    \midrule
    \multicolumn{13}{c}{\textit{LLaVA-NeXT-Video-7B-hf}} \\
    \midrule
    \texttt{Feedback (ICL)} & {2.5} & {2.0} & {3.5} & {1.78} & {0.0$^\ast$} & {0.0$^\ast$} & {0.0$^\ast$} & {0.0$^\ast$} & {0.0$^\ast$} & {0.0$^\ast$} & {0.0$^\ast$} & {0.0$^\ast$}\\
    \texttt{Realtime} & {1.85} & {2.57} & {2.71} & {2.07} & {3.01} & {3.5} & {3.89} & {2.92} & {2.75} & {3.15} & {2.78} & {2.01}\\
     \midrule
    \multicolumn{13}{c}{\textit{GPT-4.1}} \\
    \midrule
    \texttt{Feedback (ICL)} & {3.00} & {2.20} & {3.00} & {2.60} & {3.17} & {2.75} & {3.00} & {4.00} & {3.00} & {1.93} & {2.36} & {3.29}\\
    \texttt{Realtime} & {3.12} & {3.5} & {3.25} & {3.13} & {3.22} & {4.00} & {4.00} & {4.00} & {3.29} & {3.29} & {3.57} & {3.93}\\
     \midrule
    \multicolumn{13}{c}{\textit{Qwen2.5-VL-7B-Instruct}} \\
    \midrule
    \texttt{Feedback (ICL)} & {2.07} & {1.71} & {2.50} & {2.14} & {2.07} & {1.64} & {1.28} & {1.14} & {2.07} & {1.21} & {1.29} & {1.86}\\
    \texttt{Realtime} & {2.15} & {3.1} & {3.07} & {2.61} & {2.92} & {3.85} & {3.92} & {3.08} & {2.64} & {2.93} & {1.57} & {1.64}\\
    \bottomrule
    \end{tabular}
    }
    \caption{Human Evaluation scores on all 3 datasets across Key Identification Event (KEI), Patience, Conference (Coh), and Naturalness (Nat). A score of 0.0 marked with an asterisk ($^\ast$) indicates cases where the output was either in the wrong language or completely ungrammatical and unintelligible.}
    \label{tab:subjective_evaluation}
 
\end{table*}

\subsection{Additional Results}
\paragraph{Step size vs Utterance Timing}

Our approach involves incrementally feeding chunks of the video to the model, along with the previously generated commentary. The size of each chunk is determined by the step size; a larger step size results in fewer iterations within the feedback loop per sample but may introduce greater latency in the generated commentary.

We conducted experiments with various values of step size to evaluate its impact on the timing of utterances, as shown in \autoref{tab:step_size}. The results indicate that a smaller step size generates commentary that is more closely aligned with the reference prediction timing, although they also increase the complexity of our method. Based on these findings, we adopted a step size of 2 in our experiments as a compromise between performance accuracy and methodological simplicity.

\begin{table}[]
\resizebox{\columnwidth}{!}{%
\centering
    \begin{tabular}{lc c c}
    \toprule
     & \multicolumn{1}{c}{\textbf{Race (En)}} & \multicolumn{1}{c}{\textbf{Race (Ja)}} & \multicolumn{1}{c}{\textbf{Fight (Ja)}} \\ 
    \midrule
    \multicolumn{4}{c}{\textit{LLaVA-NeXT-Video-7B-hf}} \\
    \midrule
    \texttt{Feedback (ICL)} & {0.12} & {0.40} & {0.17} \\
    \texttt{Realtime} & {0.046} & {0.61} & {0.74} \\
     \midrule
    \multicolumn{4}{c}{\textit{GPT-4.1}} \\
    \midrule
    \texttt{Feedback (ICL)} & {0.55} & {0.56} & {0.97} \\ 
    \texttt{Realtime} & {0.0} & {0.0} & {0.0} \\ 
     \midrule
    \multicolumn{4}{c}{\textit{Qwen2.5-VL-7B-Instruct}} \\
    \midrule
    \texttt{Feedback (ICL)} & {0.81} & {0.93} & {0.79} \\
    \texttt{Realtime} & {0.0} & {0.0} & {0.0} \\
    \bottomrule
    \end{tabular}
    }
    \caption{Proportion of overlap in commentaries on all 3 datasets on Fixed-Interval and dynamic-interval, both using Feedback-based decoding.}
    \label{tab:overlap_commentaries}
 
\end{table}
\paragraph{Generation Statistics} We present the average number of words in the commentaries generated by the three models, averaged across all decoding strategies, as shown in \autoref{tab:llm-avg-generation-word}. The outputs produced by the multimodal large language models (MLLMs) are significantly more verbose than the reference commentaries. This indicates that, despite our method's successful integration of pause-awareness, the generated content at appropriate intervals remains relatively verbose. These findings highlight an opportunity for future work focused on guiding MLLMs towards producing more concise commentaries.

\paragraph{Contextual Similarity computed over video segments.} We show the contextual similarity of LLM generated commentary to the reference commentaries, compute over 10\% segments of the whole commentary in \autoref{fig:bertscore=bin}. Our experiments show that the commentaries generated by MLLM during the first segments are quite similar to the human-written reference summaries. However, this similarity decreases as the MLLM goes deep into the video and tends to go up in the middle or towards the end of the video.

\paragraph{Overlapping Commentaries}
Since speaking speed varies across speakers, we estimate it from the reference commentaries. For the fixed-interval decoding strategy, we prompt the LLM every 2 seconds to decide whether to wait or generate. The generated commentaries are then converted into .srt files using the estimated speaking speed. However, this naive approach can lead to overlaps in the commentary timings, as shown in \autoref{tab:overlap_commentaries}. Our dynamic-interval decoding strategy addresses this issue by tracking the estimated duration required for the speaker to narrate each commentary. It uses the elapsed time to adjust the timing (t) for the next generation, thereby reducing overlaps.

\begin{CJK}{UTF8}{min}
\begin{table}[]
\centering
\begin{tabular}{p{7cm}l}
\textbf{Car Racing (Feedback - ICL)}\\
\toprule 

00:00:00,000 --> 00:00:07,846 \\Eight racers, Shaun Akiyama, Marquis Morrill, Kurtis Nadeau, Brenton Shull, and Jordon Ordonez, compete in a thrilling race on the Monza Circuit, each driving their unique vehicles. \\\\

00:00:18,000 --> 00:00:24,102\\ Shaun Akiyana leads the pack, followed closely by Marquis Morril and Kurtis Nadau. Brenton Shul and Jordon Ordenez trail further back. \\\\

00:00:20,000 --> 00:00:24,940\\ Shaun Akyama maintains his lead, with Marquis Morril closing in quickly. Kurtis Nadel remains steady behind them. \\\\

00:00:22,000 --> 00:00:26,940\\ Marquis Morril closes in on Shaun Akiayama, while Kurtis Naden stays steady behind them, maintaining his position. \\\\


\textbf{Car Racing (Realtime)}\\
\toprule 

00:00:00,000 --> 00:00:08,616 \\
Eight racers, Shaun Akiyama in a red car, Marquis Morrill in a blue car, and Kurtis Nadeau in a yellow car, prepare for the race at the starting grid.\\\\

00:00:21,000 --> 00:00:24,565\\
As the race begins, the drivers accelerate fiercely, vying for the lead.\\\\

00:00:36,000 --> 00:00:43,427\\
Shaun Akiyana in the red car leads the pack, followed closely by Marquis Morril in the blue car and Kurtis Nadau in the yellow car.\\\\


\\
00:01:15,000 --> 00:01:20,644\\
Marquis Morrill edges closer to Shaun Akiayma, leaving Kurtis Nadel in the dust as they navigate the first turn.\\

\end{tabular}
\caption{The first four generations by Qwen2.5 using Feedback (ICL) and Realtime decoding method on Race Commentary (en). The examples show that the Feedback method, even with ICL examples, is quite verbose. The Realtime method, on the other hand, generates more concise and equidistant commentaries.}
\label{app:generation_examples}
\end{table}
\end{CJK}

\section{Conclusion}
This paper explored whether MLLMs can perform real-time video commentary generation through prompting alone, without task-specific fine-tuning. We proposed two pause-aware decoding strategies: (i) fixed-interval and (ii) dynamic interval-based decoding. We evaluated these strategies on Japanese and English datasets in the racing and fighting game domains. While automatic metrics favored the fixed-interval decoding strategy, subjective evaluations by human annotators indicated that dynamic interval-based decoding was more effective, particularly regarding Key Event Identification and pause-awareness.

Given the unreliable nature of automatic evaluation metrics in this context, our conclusions are primarily based on subjective human evaluations. Our experiments demonstrated that dynamic interval-based prompting yields commentary better aligned with human utterance timing and is perceived as more natural and pause-aware by annotators. 
Although automatic metrics revealed inconsistencies and limitations in capturing nuanced human judgments, our work represents a first step toward decoding-aware prompting for real-time language generation. In future work, we aim to explore more robust timing estimation approaches, extend our methods to broader domains, and develop improved evaluation methodologies that better correlate with human perception of timing and relevance.

\newpage
\section{Acknowledgements}

This paper is based on results obtained from a project, Programs for Bridging the gap between R\&D and the IDeal society (society 5.0) and Generating Economic and social value (BRIDGE)/Practical Global Research in the AI × Robotics Services, implemented by the Cabinet Office, Government of Japan.
This paper is also based on results obtained from AIST policy-based budget project ``R\&D on Generative AI Foundation Models for the Physical Domain.

\section{Ethical considerations and limitations}

The deployment of automated real-time commentary systems raises important ethical considerations. First, there is the potential for bias in the generated content, especially if the underlying language models were trained on data reflecting cultural or contextual biases, which could lead to disrespectful, inaccurate, or inappropriate commentary. Ensuring cultural sensitivity and fairness is critical, particularly in diverse or international audiences. Second, the use of such systems in live settings might affect the livelihoods of professional commentators or broadcasters, raising questions about the economic impact and the responsible use of automation in entertainment. Third, privacy concerns may arise if the system processes sensitive or personally identifiable information embedded within video content. Researchers and practitioners should strive to evaluate and mitigate these issues by implementing safeguards, transparency measures, and inclusive evaluation protocols to promote ethical and responsible deployment of such systems.

Although the prompt-based strategies we have proposed show promising results for pause-aware real-time video commentary, there are some limitations to consider. Firstly, the reliance on in-context prompts without fine-tuning may limit the model's ability to handle highly domain-specific or nuanced comments, especially in complex or less representative scenarios. Secondly, the dynamic interval approach relies on accurate estimation of utterance duration, which can vary considerably across languages, video genres, or real-time conditions, potentially impairing robustness. Thirdly, our evaluation focuses primarily on racing and fighting game datasets; the generalizability of these methods to other domains or content types has yet to be validated. In addition, the approach requires high-quality video data and synchronized commentary, which are not always available in practical applications.
\section{Bibliographical References}\label{sec:reference}
\bibliographystyle{lrec2026-natbib}
\bibliography{lrec2026}

\newpage
\section*{Appendix}
\begin{CJK}{UTF8}{min}
\section{Full Prompts}
\label{sec:appendix:prompts}
The prompts used by our method during decoding and initialization steps are shown in \autoref{app:inference_prompt} and \autoref{app:inference_prompt_warmup}, respectively.

\begin{table*}
\centering
\begin{tabular}{p{15.4cm}l}
\textbf{Car Racing (English)}\\
\toprule 
You are a professional commentator for car racing games. You will be provided with a video clip that represents the start of a race. Your task is to generate one sentence of commentary. \\
1) You should identify the number of players and their names, along with cars.\\
2) Ignore the background information and refrain from describing the scenery.\\
3) Initial information about the game without being too verbose.\\

\\
\textbf{Car Racing (Japanese)}\\
\toprule 

あなたはカーレースのプロの実況者です。これからレース開始時のビデオクリップが提示されます。\\
それに対して1文の日本語実況を生成してください。\\
冗長になりすぎず、レースの初期情報を伝えてください。人名や車種には言及せず「プレイヤー」や車の色を使って説明してください． \\

\\
\textbf{Fighting (Japanese)}\\
\toprule 
あなたは大乱闘スマッシュブラザーズのプロの実況者です。これから対戦開始時のビデオクリップが提示されます。\\
このシーンを1文で説明する日本語の実況を生成し視聴者を楽しませてください。\\
観客が没入できるよう驚きや感嘆句も含めてエキサイティングな実況となるよう心がけてください。話すべきことがなければ <WAIT> を出力してください。\\
\\
\end{tabular}
\caption{The prompts used by our methods during the initialization step for all three datasets.}
\label{app:inference_prompt_warmup}
\end{table*}

\begin{table*}
\centering
\begin{tabular}{p{15.4cm}l}
\textbf{Car Racing (English)}\\
\toprule 
You are a professional commentator for car racing games. You are provided with a video clip from an ongoing car racing game and commentary generated for the game so far. \\
Previous generated Commentary: \{context\} \\
Your task is to compare the given video with the previously generated commentary. \\
1) Identify if the video has any new development as compared to the already provided commentary. \\
2) Ignore the background information and refrain from describing the scenery too much. \\
3) If the state of the game as compared to the provided commentary has not changed, then generate <WAIT> \\
4) If there are new developments in the provided video, then generate 1 - 2 lines of commentary to describe it. \\

\\
\textbf{Car Racing (Japanese)}\\
\toprule 

あなたはカーレースのプロの実況者です。以下に示すのは現在進行中のレースのビデオクリップと、これまでに生成された実況です。 \\
これまでの実況: \{context\} \\
以下のルールに従って日本語実況を1〜2文生成してください： \\
1) 新たな展開があるかどうかを特定してください。 \\
2) 背景や風景の描写は避けてください \\
3) 変化がある場合は、それを説明する1文の実況を生成してください。 \\
4) 人名や車種には言及せず「プレイヤー」や車の色を使って説明してください． \\

\\
\textbf{Fighting (Japanese)}\\
\toprule 
あなたはカーレースのプロの実況者です。以下に示すのは現在進行中のレースのビデオクリップと、これまでに生成された実況です。 \\
これまでの実況: \{context\} \\
ビデオに新たな展開があるかどうかを比較・分析し、以下のルールに従って日本語実況を生成してください： \\
1) 新たな展開があるかどうかを特定してください。 \\
2) 状況に変化がなければ <WAIT> を出力してください。 \\
3) 明確な変化があれば、それを説明する1文の実況を生成してください。 \\
4) 人名や車種には言及せず「プレイヤー」や車の色を使って説明してください\\
\\
\end{tabular}
\caption{The prompts used by our methods during inference on all three datasets.}
\label{app:inference_prompt}
\end{table*}

\section{Human Evaluation}

\label{appendix:HE-instructions}
The guidelines given to the Human Annotators for evaluating the generated commentary are shown in \autoref{app:guidlines-HE}.

\begin{table*}
\centering
\begin{tabular}{p{15.4cm}l}

\textbf{General Instructions}\\
\toprule

Each score should be evaluated independently (e.g., a commentary may be fluent but still fail to identify key events). \\
A score of 0 should only be used if the commentary is unreadable or in the wrong language. \\
When the commentary includes multiple sentences, evaluate based on the overall impression and frequency of issues.\\
Recommended Evaluation Workflow (Example)\\
Play or read the generated commentary alongside the video or reference.\\
Assign a score (0–5) for each of the four criteria.\\
Optionally add comments (especially if any score is 3 or below) to assist in qualitative analysis.\\
\\

\textbf{1. KEI (Key Event Identification)}\\
\toprule 

Evaluate whether the commentary correctly identifies important events (e.g., actions, transitions, score changes).\\
0: Unintelligible or not in the target language.\\
1: No events are mentioned, or all mentions are clearly incorrect.\\
2: Events are mentioned but mostly inaccurate or severely mistimed.\\
3: Some key events are accurately identified, but others are missed or incorrectly described.\\
4: Most key events are correctly described, with only minor inaccuracies.\\
5: All major events are accurately and appropriately described.\\

\\
\textbf{2. Pause-awareness (Timing Appropriateness)}\\
\toprule

Evaluate whether the model refrains from speaking unnecessarily and speaks at appropriate times. \\
0: Unintelligible or not in the target language. \\
1: Continuously speaking with no pauses; timing is too fast to follow. \\
2: Overly verbose or too silent, significantly disrupting the viewing experience. \\
3: Slightly too fast or too slow, but still followable. \\
4: Overall pacing is natural, with only minor issues. \\
5: Excellent balance between silence and speech; timing feels natural and appropriate. \\

\\
\textbf{3. Coherence (Logical Consistency)}\\
\toprule 
Evaluate whether the commentary is logically consistent and maintains a clear narrative flow. \\
0: Unintelligible or not in the target language. \\
1: Individual utterances make sense, but logical flow is broken or contradictory. \\
2: Weak narrative connection with parts that cause confusion or inconsistency. \\
3: Mostly coherent with minor awkward transitions or inconsistencies.\\
4: Clear and consistent overall flow, with only slight unnaturalness.\\
5: Fully coherent and logically well-structured throughout.\\
\\

\textbf{4. Naturalness (Linguistic Fluency)}\\
\toprule 

Evaluate whether the language is fluent and sounds natural in the given language (grammar, phrasing, word choice).\\
0: Unintelligible or not in the target language.\\
1: Grammatical or unnatural phrases dominate; strong sense of artificiality.\\
2: Grammatically correct but awkward or mechanical phrasing.\\
3: Mostly fluent, with a few unnatural or stiff expressions.\\
4: Fluent and natural overall, with only slight awkwardness.\\
5: Highly fluent and natural; indistinguishable from human-spoken commentary.\\
\\
\end{tabular}
\caption{Evaluation Dimensions and Scoring Criteria for Human Evaluation}
\label{app:guidlines-HE}
\end{table*}

\end{CJK}
\end{document}